\documentclass[english,11pt]{article}
\usepackage{geometry}
\usepackage{float}
\usepackage{pgf,tikz}
\usepackage{mathrsfs}
\usetikzlibrary{arrows}
\usepackage{ulem}
\usepackage{comment}
\usepackage{mathrsfs,bm}
\usepackage{amsmath}
\usepackage{amssymb}
\usepackage{fancyhdr}
\usepackage{epsfig}
\usepackage{amsthm}
\usepackage{mathrsfs}
\usepackage{amsfonts}
\usepackage{mathrsfs}
\usepackage{enumerate}
\usepackage{subfigure}
\usepackage{setspace}
\usepackage{mathtools}
\usepackage{natbib}
\usepackage{tikz}
\usepackage{tikz-3dplot} 
\usepackage{pgfplots}
\newcommand{\indep}{\perp \!\!\! \perp}
\usepackage{amsmath,amssymb}
\usepackage{caption}
\usepackage{color}
\usepackage{dcolumn}
\usepackage{bm}
\usepackage{tikz}
\usepackage{tikz-3dplot} 
\usepackage{pgfplots}
\pgfplotsset{compat=1.11}
\usepackage{algorithm}
\usepackage[noend]{algpseudocode}
\usepackage{tikz}
\usetikzlibrary{bayesnet}
\usepackage{amssymb}
\usepackage{caption}
\usepackage{color}
\usepackage{dcolumn}
\usepackage{etoolbox}

\usepackage{amsmath}
\usepackage[noend]{algpseudocode}

\usepackage[utf8]{inputenc}

\usepackage[final]{hyperref} 
\usepackage{setspace}

\newcommand{\doc}[1]{\operatorname{do}{(#1)}}

\newcommand{\qref}[1]{Eq.~(\ref{#1})}

\newcommand{\cE}[2]{\mathbb{E}\left[#1\middle|#2\right]}
\newcommand{\cV}[2]{\mathbb{V}\left[#1\middle|#2\right]}
\newcommand{\X}{\mathbf{X}}

\newcommand{\bO}{\mathbf{O}}
\newcommand{\Q}{\mathbf{Q}}
\newcommand{\I}{\mathbf{I}}
\newcommand{\q}{\mathbf{q}}
\newcommand{\bi}{\mathbf{i}}
\newcommand{\calD}{\mathcal{D}}
\newcommand{\x}{\mathbf{x}}
\newcommand{\calG}{\mathcal{G}}
\newcommand{\G}{\mathbf{G}}
\newcommand{\A}{\mathbf{A}}
\newcommand{\bb}{\mathbf{b}}
\newcommand{\bq}{\mathbf{q}}
\newcommand{\parents}[2]{\operatorname{Pa}_{#1}(#2)}
\newcommand{\btheta}{\boldsymbol{\theta}}

\usepackage{breqn}

\title{Bayesian Model Averaging for Data Driven Decision Making when Causality is Partially Known}
\author{Marios Papamichalis \thanks{Statistics Department, Purdue University, mpapamic@purdue.edu.}, {Abhishek Ray\thanks{School of Business, George Mason University, aray8@gmu.edu.}}, {Ilias Bilionis\thanks{Department of Mechanical Engineering, Purdue University,ibilion@purdue.edu.}}, {Karthik Kannan\thanks{Krannert School of Management, Purdue University, kkarthik@purdue.edu.}}, {Rajiv Krishnamurthy\thanks{Independent Researcher, rajivjk@acm.org.}}}

\begin{document}

\maketitle

\begin{abstract}

Probabilistic machine learning models are often insufficient to help with decisions on interventions because those models find correlations - not causal relationships. If  observational data is only available and experimentations are infeasible, the correct approach to study the impact of an intervention  is to invoke Pearl’s causality framework. Even that framework assumes that the underlying causal graph is known, which is seldom the case in practice. When the causal structure is not known, one may use out-of-the-box algorithms to find causal dependencies from observational data. However, there exists no method that also accounts for the decision-maker's prior knowledge when developing the causal structure either.  The objective of this paper is to develop rational approaches for making decisions from observational data in the presence of causal graph uncertainty and prior knowledge from the decision-maker.  We use ensemble methods like Bayesian Model Averaging (BMA) to infer set of causal graphs that can represent the data generation process. We provide decisions by computing the expected value and risk of potential interventions explicitly.  We demonstrate our approach by applying them in different example contexts.


\end{abstract}

\newpage

\section{Introduction}

\doublespacing

A key benefit of Big Data is that it allows firms to perform online controlled experiments at massive scale \citep{kohavi2013online}. Scale in this context refers to the number of experiments a firm runs; some estimates suggest that internet firms conduct over 10,000 experiments annually \citep{kohavi2017surprising}. Each of these experiments allows decision makers to make precise, causal connections between product changes and key business metrics (such as revenue, engagement and cost). This allows decision makers to make decisions with confidence and certainty. With the volume of experimentation being such, it is no surprise that internet firms innovate at an unprecedented pace. Not surprisingly, IS researchers  use online experimentation extensively \citep[e.g.,][]{tianshu21}.

While online experimentation has emerged as the default go-to for product related decisions in the internet industry, there exists a class of product decisions for which online experimentation is either too costly or impossible even when large amounts of data is available. We motivate this class of decisions with an example -- one way in which firms make their products better is by making it more efficient with respect to platform characteristics, e.g., more efficient apps considering smartphone resources such as battery, bandwidth, memory etc. These are critical factors known to be associated with app installs and usage (Ickin et.a 2017) -- a key consideration from a revenue standpoint. While ideally firms may want to optimize all these factors at once, in practice they have to prioritize. This is because making these optimizations can be very costly in terms of engineering R\&D. Firms usually measure various parameters within their apps in the form of time series, panel or cross-sectional data, so it is natural to consider online experiments to guide this prioritization. However, here lies the predicament - one cannot run an ``experiment” to measure the business impact of one optimization and contrast it with the business impact of another to decide between the two -- simply put, optimizations once made (at considerable cost) should be made available to all users. In such circumstances when experiments are moot, practitioners either rely on their intuition or resort to some correlational analysis, as provided by common machine learning or probabilistic algorithms, to make a decision. Both these approaches are risky -- in the case of the latter incorrect assumptions about the underlying causal structure can lead to incorrect insights.\footnote{For example, if a hotel observes high correlation between the occupancy rate and price charged for the room, a manager may incorrectly infer, without considering the demand, that charging higher prices leads to higher occupancy.}

Drawing motivation from scenarios described above, we focus on developing a robust, data driven procedure for decision-making that incorporates partial knowledge of causal information. More precisely, our major contributions are as follows: first, we provide a novel algorithm that combines structure learning and MH-MCMC approaches to generate a class of causal graphical models from high-dimensional data. This class of causal networks incorporates hybrid information both from the decision maker’s prior knowledge and the associated data available - ideal for real world scenarios described previously. Second, using these class of models we build a decision-making framework that provides insights on how practitioners should act if a given graphical model representing the posterior distribution is not trustworthy. Specifically, our proposed framework overcomes common problems during application, e.g., data misspecification, by quantifying the prior beliefs of the practitioner and using subsets of graphical models that fit them. Finally, we demonstrate the benefits from our methodology by generating results using example scenarios.


The rest of the paper proceeds as follows.  Section \ref{litReview} describes the literature including Bayesian networks, Pearl's model and Bayesian Model Averaging. Then, in Section 3, we focus on our main purpose of this paper which is how to use and apply current tools on causal inference to see how much information can we use in order to decide better.  In section \ref{Sim}, we apply our method for network data analysis that provides actionable managerial direction. Finally, in section 5, we present with more details the future work involving overlapping research areas. 

\section{Literature Review}\label{litReview}
Most practical decision-making contexts (e.g., manufacturing, portfolio selection) use a classical statistical approach.  Their models assume simplified representations of causality for estimation purposes.  The associated risk with those models is attributed to the inherent randomness in the world. Therefore, they employ probabilistic expressions to represent the randomness.   Thus, the classical evidential paradigm captures aleatory uncertainty.  However, Bayesian statisticians attribute the uncertainty to the modeler's lack of complete knowledge or epistemic uncertainty.  Even they begin with the causal model.  However, the parameters of the model are estimated by assuming a prior  \citep{azoury1985bayes,berger2013statistical,robert2007bayesian}. Naturally, the estimation is highly dependent on the assumption regarding the priors.  This is a big source of concern for frequentists.  The differences between the two schools of thought regarding statistical theory is well known.  Even beyond that, there are additional issues are pertinent to decision-making. 

\subsection{Practical Constraints in Using Decision-Making Models}
In many companies and also in much of the prior work, the estimation task is  conducted separately from the optimization  or decision-making task.  For example, large retailers have a demand forecasting team that is different from the inventory optimizing team.  Even though the forecasting demand is typically  focused on minimizing the statistical errors, the company as a whole should be concerned about mitigating the error at the optimization stage.  Using a Bayesian analysis, \citet{chu2008solving,liyanage2005practical,ramamurthy2012inventory} provably demonstrate the value of  operational statistic  useful when jointly considering both estimation and optimization tasks.  Specifically, they demonstrate that operational statistic can be used with data in a news-vendor context.  So, there is a clear need for considering the estimation and optimization tasks jointly.

Across the aforementioned research, a common theme is the assumption that the causal relationships are known.  While such an assumption is often made, in reality, there are many situations where the causal relationships are not fully known even to the decision-maker.  In the big data world, this issue becomes even more salient.  As \cite{feng2018research} note: ``though theoretically more information leads to better forecasts, the challenge, however, comes from dealing with the increased number of variables and their ambiguous relationships.'' The practical need therefore is to synthesize robust decision-making methodologies keeping the realities and demands under consideration. 
\subsection{Causal Inference}
A consistent view is that it is most effective to study the causal impact is through experimentation.  Given the easiness with which experiments can be conducted in the Internet world, not surprisingly, Internet companies conduct thousands of experiments before deciding on the interventions. Existing research has examined various instances where firms use experiments to investigate causal implications of changes in app features on app adoption decisions \citep{sun2019motivating}, customer conversions \citep{huang2019word} and offline group formations \citep{sun2019mobile}. More recently, research has also looked at causal implications of digital technologies on offline activity, e.g., dating apps and matching outcomes \citep{bapna2016one}, app adoption and dating app usage \citep{jung2019love}. However, the luxury of experimentation does not always prevail, sometimes not even for the Internet firms.  Our interaction with an app company indicated that the firm was constrained by engineering team resources to conduct some of the  experiments.  In those situations, all the aforementioned issues arise and one has to consider alternatives regarding how to use observational data for inference.  There are differing viewpoints in that regard.  Rubin's framework assumes that there is ``no causation without manipulation."  However, Pearl's framework assumes that, if the data is associated with a causal graph of variables (including the intervention and the outcome) and the graph  blocks the backdoor paths between the intervention and the outcome, one may be able to study the causal impact using the data.  This leads us to identifying the causal graphs from observational data.

Since the 1990s, conditional independence associations in data have been used to recreate the underlying causal structure. The PC algorithm is a popular constraint-based algorithm that discovers conditional independence. The PC algorithm claims that there is no confounder and that the causal information found is asymptotically correct. Since it can accommodate a broad range of data distributions and causal relationships, PC approach is broadly applicable. However, since the PC algorithm outputs independent equivalence classes, or a series of causal structures that are constrained by the same conditional independence conditions, it does not always provide full causal knowledge. Causal discovery, a more advanced way of discovering causal evidence by studying solely observational evidence, has also gained popularity \citep{Spirtes}. In genomics, genetics, epidemiology, space mechanics, clinical medicine, psychiatry, and many other fields, causal discovery has also been used. For example, in gene micro array data inference algorithms have been used to discover network structure \citep{Murphy, Friedman,Spirtes}. Recently, Bayesian Networks were introduced to a modern biological paradigm and shown to outperform previous regression-based approaches in revealing how data from brain electrophysiology are used to build flow networks for neural information \citep{Smith}. Such a versatile tool capable of defining the diverse interactions involved in gene and neuronal control may also provide a useful new approach for interpreting data in SBE contexts. Bayes Networks are a powerful tool for graphical representation and factoring of combined probability distributions. As a consequence, when it comes to networks, they are better at detecting results and they behave better for variables that have several nuanced interrelationships \citep{Koller}. Discrete Bayesian networks, for example, are powerful methods for inferring network structure from field data \citep{Heckerman}, and is a fast developing area of research.

Since learning Bayesian Networks from data is NP hard,  any discovered graph is only an approximate representation.   Bayesian Model Averaging \citep{yao2018using,fragoso2018bayesian,claeskens2008model,hoeting1999bayesian} averages out the network structures and the coefficients to obtain the joint distribution that captures the most likely existing causal relationships.  Specifically, Bayesian model averaging is an ensemble  that is simply averaged over the separate models, where  each is weighted by its marginal posterior probability.

\subsection{Bayesian Networks}

Generally, a network is defined as a tuple $G = (V, E)$, where $V$ denotes the set of nodes, and $E$ the set of edges $E \in V \times V$. $A_G$ denotes the adjacency matrix of $G$. Suppose we use $n$ to denote $\mid V \mid$. A simple network is a network with at most one edge between any pair of nodes and without any self-loop. Further, $G$ is directed if the corresponding adjacency matrix is asymmetric. A set $\mathcal{G}$ of networks is the space of all such adjacency matrices collected.

Bayesian networks are an easy way to describe and simulate the interactions between a wide number of variables. The conditional independence/correlation relationships between the variables can be abstracted from the specifics of their parametric types using a matrix. Crucially, Bayes nets can be used to provide a statistical foundation for debates about causality. This is because these networks have a very efficient way to graphically depict and factor joint probability distributions, making them perfect for classification. The nodes (vertices) of a Bayesian network represent random variables from the population of interest, while the arcs (edges) represent the random variables' immediate dependencies. For a Bayesian network with $n$ nodes $X_1, X_2, \dots , X_n$ the full joint probability density is defined as:
\begin{equation}
p(x_1, x_2, \dots , x_n) = \prod_{i=1}^n p(x_i |parents(X_i)) 
\end{equation}
where $p(x_1, x_2, \dots , x_n)$ is an abbreviation for $p(X_1 = x_1 \wedge
\dots \wedge X_n = x_n)$.
They identify automatically the joint distribution of the data and find all the patterns of connections to explain (or predict) the target variable. Bayesian networks are better at finding effects when it comes to networks because they behave better with variables that have many subtle interrelationships \citep{Koller}.
In terms of analysis, we differentiate between cases where a vector X naturally takes the value $x$ and cases where there is explicit \textit{interference} to assign $X = x$ by denoting the latter $do(X = x)$, called do-calculus. In other words, do-calculus implies causation, in observational studies. So $P(Y = y \mid X=x)$ is the probability that $Y = y$ conditional on finding $X = x$, while $P(Y = y \mid do(X = x))$ is the probability that $Y = y$ when there is explicit intervention to make $X = x$.  Similarly $P(Y = y \mid do(X = x), Z = z)$ is to denote the conditional probability of $Y = y$, given $Z = z$, created by the intervention $do(X = x)$.

In a causal sense, two or more variables are believed to be associated if their values shift such that when one variable's value rises or falls, so does the value of the other variable (although the directionality might be opposite). This pattern would mean that one event happens as a result of the existence of the other, meaning that the two occurrences have a causal connection. This implies that the change in one variable is the source of the change in the other variable's values. Trigger and impact is another term for this. The uniform values of causal coefficients, such as correlation coefficients, are used, which vary in -1 to 1.

More information about the differences between experimental designs and causal dependencies through Pearl's framework are described thoroughly in the appendix. Moreover, meticulously, the algorithm we use to catch those dependencies (PC algorithm) is presented. 

\subsection{Bayesian Model Averaging in Causal inference}
Bayesian Model Averaging \citep{yao2018using,fragoso2018bayesian,claeskens2008model,hoeting1999bayesian} averages out the network structures and the coefficients, created through an MH-MCMC algorithm, to get the joint distribution which captures the most likely existing causal relationships. 
BMA is particularly useful where several competing models remain viable and the goal is prediction or parameter estimation. In this situation, models cause more uncertainty than clarity; they aren't directly relevant, but model selection has an effect on prediction and estimation. BMA clears up the ambiguity by including the best forecasts or parameter projections depending on the candidate models' beliefs. When one model dominates the others, however, BMA is less useful. Another situation where BMA is helpful is where control variables are unpredictable. BMA can be helpful where a researcher has to evaluate the data in favour of two or three conflicting measures that are likely associated. Researchers may also use BMA to perform more consistently robustness tests of their predictions than is feasible with a frequentist approach. Finally, BMA supports researchers who want to approximate the results of a wide range of possible predictors of a significant dependent variable. In all of these cases, recent methodological advancements have increased the usefulness of BMA.

Typical model selection methods use AIC or BIC as selection criteria, apart from the Bayesian approach. In the BMA context  model selection is typically driven with the estimation of causal effects in mind \citep{ertefaie2018variable,vansteelandt2012model}. For instance,  \cite{ertefaie2018variable} proposes a model selection approach focusing on both the outcome and treatment assignment models and using a penalized target function. The proposed approach makes confounder selection easier in high-dimensional environments, and it has been shown to be very successful in revealing causal dependencies in observational results. Overall, BMA picks the best candidate model asymptotically \citep{yao2018using,fragoso2018bayesian,claeskens2008model,hoeting1999bayesian}. If all of the candidate models are generative, the Bayesian approach is to merely average them, evaluating them according to their marginal posterior likelihood. Briefly, for the three following cases: first, $\mathcal{M}$-closed implies the true data generating model ($M_k \in M$)
is unknown to researchers; second, $\mathcal{M}$-complete implies the true model exists and is out of scope of $\mathcal{M}$;
third, $\mathcal{M}$-open implies that the true model $\mathcal{M}_t \notin \mathcal{M}$ and is impossible to specify -- BMA is a good way to pick the right model asymptotically for the $\mathcal{M}$-closed case. In that case, the weight of the possibly true model is expected to converge to 1, making the weighted average of models (BMA) into a collection of the single best one in the large-sample-limit of results. In $\mathcal{M}$-open and $\mathcal{M}$-complete cases, BMA will asymptotically choose the variant on the list with the smallest Kullback-Leibler (KL) divergence from the others. Based on the evidence available, this is the nearest model to a true model.


\section{Methodology}\label{method}


Consider a representative firm that is interested in enhancing some quality variables $\Q$ of its product. The random variable $\Q$ may be single-valued, e.g., a software company may be interested in improving user engagement -- measured as the amount of time consumers spend using its product, or multi-valued. In any case, we introduce a value function, $v(\q)$, that ranks quality tuples as per the decision-maker's preference, i.e., the decision-maker prefers $\q_1$ to $\q_2$ if and only if $v(\q_1) > v(\q_2)$.
The quality $Q$ may be improved by intervening on some parameters related to the product. 
We call these parameters intervention variables and we denote them by $\I$. 
For example, in the context of software development, intervention variables could be the ``time to cold start,'' and ``battery power consumption rate,'' to name a few. 
Setting the intervention variable $\I$ to a specific value $\bi$ requires engineering effort and, therefore, has a cost $c(\bi)$.
It is important that the cost is measured in the same units as value, so that the net value of an intervention $\bi$ resulting in quality $\q$ is $v(\q)-c(\bi)$.
Of course, if the cost of all interventions is the same, then it can be eliminated from the analysis.
We wish to study the decision of how the firm should prioritize which intervention to make to improve quality.

The decision to improve quality presents a trade-off -- modify a select set of features to maximize quality while minimizing risk from such modifications. 
This trade-off can be resolved through the introduction of a subjective utility function, but in this work we focus on characterizing the Pareto front of maximum-quality vs minimum-risk decisions because firms typically want to know what is possible before balancing gain and risk through a utility function.
The naïve approach to this problem considers maximizing the expectation and minimizing the variance of the net value $v(\Q)-c(\I)$ conditioned on an intervention $\I=\bi$.
Mathematically, the problem becomes ranking interventions according to the two-objective optimization problem:
\begin{equation}\label{eqn:naive}
    \begin{split}
        &\max_{\bi} \cE{v(\Q)-c(\I)}{\I=\bi}=\max_\bi \left\{\cE{v(\Q)}{\I=\bi}-c(\bi)\right\},\\
        &\min_{\bi} \cV{v(\Q)-c(\I)}{\I=\bi}=\min_\bi\cV{v(\Q)}{\I=\bi},
    \end{split}
\end{equation}
where $\mathbb{E}[\cdot|\cdot]$ and $\mathbb{V}[\cdot|\cdot]$ are the conditional expectation and variance operators, respectively.
In most practical situations, this approach to decision-making fails to resolve subjunctive conditionals, commonly called the Newcomb Paradox. That is, decision-makers consider conditional, also known as evidential, probabilities to be the same as causal probabilities.
However, as argued in \cite{gibbard1978counterfactuals}, evidential and causal probabilities are not the same.

Causal decision theory recommends using counterfactual probabilities for measuring impact of different interventions on the outcome. 
That is, instead of \qref{eqn:naive}, one solves the two-objective optimization problem:
\begin{equation}\label{eqn:causal}
    \begin{split}
        &\max_{\bi} \cE{v(\Q)-c(\I)}{\doc{\I=\bi}} =\max_\bi \left\{\cE{v(\Q)}{\doc{\I=\bi})}-c(\bi)\right\},\\
        &\min_{\bi} \cV{v(\Q)-c(\I)}{\doc{\I=\bi}}=\min_\bi\cV{v(\Q)}{\doc{\I=\bi}},
    \end{split}
\end{equation}
where one uses do-calculus instead.
This program is relatively simple to carry out when two conditions hold.
First, the causal structure must be known and, second, the causal probabilities $p(\Q=\q|\doc{\I=\bi})$ must be identifiable from observational data. 
In this paper, we develop a methodology that allows one to carry out the aforementioned decision-making program when these two assumptions do not hold.
Specifically, in Sec.~\ref{sec:causal_UQ} we show how one's uncertainty about the causal graph can be quantified in the first place. The variability of the uncertainty comes either from the prior information on the data, in order to perform causal inference, or from the subjective prior beliefs about the outcome of the practitioner. Furthermore, in Sec.~\ref{sec:dm_with_uncertainty_graphs} we demonstrate how one can make decisions even when uncertain about the causal graph and when the causal effect of particular interventions is unidentifiable from observational data.

\subsection{Quantifying Causal Structure Uncertainty}
\label{sec:causal_UQ}
First, we define a prior and the practitioner beliefs over the space of causal graphs. Then, we connect to the observed data by a model that does Bayesian linear regression conditioned on the causal structure.

\subsubsection{Quantifying Prior Causal Structure Uncertainty}
\label{sec:prior_uncertainty}
Suppose that we have $d$ different variables which we denote collectively by $\X = \{X_1,\dots,X_d\}$.
The quality variables $\Q$ and the intervention variables $\I$ are disjoint subsets of $\X$.
We refer to any variables that are in $\X$ but not in $\Q$ or $\I$ as the \textit{other} variables $\bO$.
In this work, we assume that we can observe all variables in $\X$, i.e., there are no hidden variables that are not observed.
The case with hidden variables requires problem-specific human intuition and it is beyond the scope of this paper. 

We represent the causal relationships between all variables in $\X$ using a directed acyclic graph (DAG) $G$.
The DAG $\G$ has $d$ nodes, one for each variable $X_j$ in $\X$.
Furthermore, $\G$ can be encoded as a $d\times d$ binary matrix.
Specifically, if $G_{ij} = 1$, then there is an edge from variable $X_i$ to variable $X_j$ and we say that $X_i$ is a direct cause of $X_j$.
Similarly, if $G_{ij}=0$, then we say that there is no direct causal link from $X_i$ to $X_j$.
The graph $\G$ is acyclic in the sense that for any node $X_i$ there is no directed path in $\G$ that starts and ends at $X_i$, i.e., the graph $\G$ does not have any cyclic directed paths.
We will be using the notation $\calG_d$ for the set of DAGs with $d$ nodes.

Before we see any data, we need to specify our beliefs about $\G$.
Essentially, we need to come up with a probability measure on $\calG_d$ that represents our state of knowledge about the causal structure.
Note that we are not after the most general characterization of this prior probability measure.
Instead, we are interested in a representation that one can easily elicit by asking simple questions.
To this end, we restrict our attention to a prior probability density over $\G$ of the form:
\begin{equation}\label{eqn:G_prior}
    p(\G|\A) \propto 1_{\calG_d}(\G)\prod_{i\not=j}p(G_{ij}|A_{ij}),
\end{equation}
where $1_B(\cdot)$ is the indicator function of a set $B$, the product is over all pairs of variables, and $\A$ is a $d \times d$ matrix with each element $A_{ij}$ taking values in $[0,1]$.
We interpret $A_{ij}$ as the prior probability that there exists a direct causal link between $X_i$ and $X_j$, i.e.,
$$
    p(G_{ij}=1|A_{ij}) = A_{ij}.
$$
In words, if $A_{ij}=1$, then we are absolutely sure that there is a direct causal link between $X_i$ and $X_j$; if $A_{ij} = 0$, then we are sure that there is no direct causal link between the two variables; and if $0 < A_{ij} < 1$, then we are uncertain about the existence of the direct causal link.
Since $G_{ij}$ is either zero or one, we can write the prior probability of any possible value of $G_{ij}$ as:
\begin{equation}\label{eqn:prior_G_ij}
    p(G_{ij}|A_{ij}) = A_{ij}^{G_{ij}}(1-A_{ij})^{1-G_{ij}}.
\end{equation}

\subsubsection{The Data Likelihood Conditioned on a Causal Structure}
\label{sec:like}
Assume that we have observed $N$ independent realizations of $\X$ denoted by:
\begin{equation}
    \calD = \{\x_1,\dots,\x_N\},
\end{equation}
where
$$
\x_n = (x_{n1},\dots,x_{nd}).
$$
In this section, we derive an expression for the data likelihood $p(\calD|\G)$ conditioned on a causal structure represented by a DAG $\G$.

First, we need to demonstrate how $\G$ induces the so-called \textit{structural equations} that connect all the variables in $\X$.
Let ${\X}_{-i}$ be all the variables in $\X$ except $X_i$ and $\parents{\G,i}{{\X}_{-i}}$ the subset of ${\X}_{-i}$ containing the variables that are direct causes of the variable $X_i$, i.e., the so-called parents of $X_i$:
\begin{equation}\label{eqn:parents_def}
    \parents{\G,i}{\X_{-i}} = \left\{X_j \in \X_{-i}: G_{ji} = 1\right\}.
\end{equation}
Note that it is possible that some variables have no parents under $\G$.
If $X_i$ is a variable with no parents, then $\parents{\G,i}{\X_{-i}}$ is the empty set $\emptyset$.

Now take a variable $X_i$ that is not a root cause.
We assume that $X_i$ takes its values through a mechanism of the form:
\begin{equation} \label{eqn:structural_equation_Xi}
    X_i = f_i(\parents{\G,i}{{\X}_{-i}}, U_i),
\end{equation}
where $U_i$ is a random variable that is independent of $\X$ and $f_i(\cdot)$ is a function.
Note than when $X_i$ has no parents, i.e., when $\parents{\G,i}{\X_{-i}}=\emptyset$, we interpret $f_i(\cdot)$ as a function of $U_i$ only.
We call Eq.~(\ref{eqn:structural_equation_Xi}) the \textit{structural equation} for $X_i$.

In what follows, we assume that $X_i$ is a scalar so that we keep the notation simple.
The model can easily be extended to the case of a discrete or vector-valued $X_i$ like in \cite{heckerman1995learning} for the discrete case describing continuous datasets as in \cite{geiger1994learning}. The general case, with both discrete and continuous datasets are presented in \cite{heckerman2013learning}
To proceed with the definition of the data likelihood, one needs to select a parameterized model for the structural equations.
The simplest such model is a linear structural equation with Gaussian noise, i.e.,
\begin{equation}\label{eqn:structrual_equation_Xi_model}
    X_i = b_{i0} + \bb_{\G,i}^T\parents{\G}{\X_{-i}} + \sigma_i U_i,
\end{equation}
where we think of $\parents{G}{\X_{-i}}$ as a vector, $b_{i0}$ is a bias, $\sigma_i$ is positive, $U_i$ is a standard normal random variable independent of $\X$, and $\bb_{\G,i}$ is a vector with the same dimension as $\parents{G}{\X_{-i}}$.
Note that if $X_i$ has no parents, then there are no parameters, i.e., $\bb_{G,i}$ is the empty set $\emptyset$.
Furthermore, in this case we define $\emptyset^T\emptyset\equiv 0$ so that Eq.~(\ref{eqn:structrual_equation_Xi_model}) remains valid.

The parameters associated with variable $X_i$ are:
\begin{equation}\label{eqn:theta_i}
    \btheta_{\G,i} = \{b_{0i}, \bb_{\G,i}, \sigma_i\}.
\end{equation}
We can interpret the structural equation ~(\ref{eqn:structrual_equation_Xi_model}) probabilistically:
\begin{equation}
    p(x_{ni}|\parents{\G,i}{\x_{n,-i}},\btheta_{\G,i}) = N(x_{ni}|b_{i0} + \bb_{\G,i}^T\parents{\G}{\x_{n,-i}}, \sigma_i^2),
\end{equation}
where $p(y|I)$ stands for the conditional PDF of a random variable $Y$ given $I$ evaluated at $Y=y$, and $N(y|\mu,\sigma^2)$ is the PDF of a Gaussian with mean $\mu$ and variance $\sigma^2$ evaluated at $y$.
Now, denote all the model parameters collectively by:
\begin{equation}\label{eqn:theta}
    \btheta_G = \{\btheta_{\G,1},\dots,\btheta_{\G,d}\}.
\end{equation}
We can write the likelihood of a single observation as:
\begin{equation}\label{eqn:single_pt_likelihood}
    p(\x_n|\btheta_{\G}, \G) = \prod_{i=1}^dp(x_{ni} \mid \parents{\G}{\x_{n,-i}},\btheta_{\G,i}).
\end{equation}
The likelihood of all the data is:
\begin{equation}\label{eqn:like_theta}
    p(\calD|\btheta_\G, \G) = \prod_{n=1}^Np(\x_n|\btheta_{\G},\G).
\end{equation}
The posterior of the parameters given the causal structure $\G$ is:
\begin{equation}\label{eqn:post_theta}
    p(\btheta_\G | \calD, \G) \propto p(\calD|\btheta_\G, \G) p(\btheta_\G |\G),
\end{equation}
where $p(\btheta_\G |\G)$ is the prior.
We choose the prior so that the parameters $\theta_{\G,i}$ are independent conditional to the causal structure $\G$.
Furthermore, we pick the $i$-specific prior $p(\btheta_{\G,i}|\G)$ to be the conjugate prior as in table 1. For both discrete and continuous datasets, we formulate the prior, posterior distribution and predictive posterior distribution as in \cite{heckerman2013learning}.
With this choice the posterior in Eq.~(\ref{eqn:post_theta}) conveniently factorizes over $i$ and each factor is analytically available.
As a result, it is possible to integrate $\btheta_\G$ out and get the \textit{marginal likelihood}:
\begin{equation}\label{eqn:marginal_like_G}
    p(\calD|\G) = \int p(\calD|\btheta_\G, \G)p(\btheta_{\G}|\G)d\btheta_\G.
\end{equation}
This is just the product of the normalization constants of all the $i$-specific posteriors. As we see in \ref{sec:posterior_G}, the marginal likelihood is essential for characterizing our posterior state of knowledge about the causal structure.

\begin{table}
\centering
\fbox{%
\begin{tabular}{| l  l |}
     \hline
     Type of variables & Formula   \\
     \hline
     Continuous Prior on $\btheta_{\G,i}$ & Normal-Inverse-Wishart \\
     \hline
     Continuous Posterior of $\btheta_{G,i}$ & Normal-Inverse-Wishart \\
     \hline
     Discrete Prior on $\btheta_{\G,i}$ &  Dirichlet distribution\\
     \hline
     Discrete Posterior of $\btheta_{G,i}$ &  Dirichlet-multinomial distribution \\
     \hline
\end{tabular}}
\caption{Formulas for Prior and Posterior distribution for both discrete and continuous cases. In discrete regime Dirichlet distribution is applied as a conjugate prior to multinomial distribution and in continuous case Normal-Inverse-Wishart distribution is used as a conjugate prior to transformed to Gaussian data, continuous data. Notation is borrowed from \cite{heckerman1995learning} and \cite{geiger1994learning} accordingly. Further information e.g. posterior predictive distributions could be found there.
}
\end{table}
Finally, let us mention how one can evaluate the joint probability density of all variables conditioned on the data and the causal structure, i.e., the so-called \textit{predictive} joint probability density.
It is:
\begin{equation}\label{eqn:predictive}
    p(\x | \calD, \G) = \int p(\x | \btheta_\G, \G) p(\btheta_\G|\calD, \G)d\btheta_\G.
\end{equation}
The term $p(\x | \btheta_\G, \G)$ is given by the data likelihood of~\qref{eqn:like_theta}.
Like before, the integral factorizes over each $i$-specific term, and it is possible to do analytically. This is the joint PDF used for the estimation of causal effects conditioned on the causal structure in later sections.\\
Relative plausibilities of other $\btheta_{\G,i}$ values may be found by comparing the likelihoods of those other values with the likelihood of $\hat{\btheta}_{\G,i}$.
\begin{equation}
 R(\theta )={\frac {{\mathcal {L}}(\btheta_{\G,i} \mid x)}{{\mathcal {L}}(\hat{\btheta}_{\G,i}\mid x)}}.
\end{equation}
Thus, the relative likelihood is the likelihood ratio (discussed above) with the fixed denominator ${\displaystyle {\mathcal {L}}({\hat{\btheta}_{\G,i}})}$. This corresponds to standardizing the likelihood to have a maximum of 1.

\subsubsection{The Posterior Causal Structure Uncertainty}
\label{sec:posterior_G}



We have defined the prior over the causal structures $p(\G|\A)$, see \qref{eqn:G_prior}, and the marginal likelihood of the data given a causal structure $p(\calD|\G)$, see \qref{eqn:marginal_like_G}.
Our posterior uncertainty about causal structure $\G$ is neatly characterized via Bayes' rule:
\begin{equation}
p(\G|\calD,\A) \propto p(\calD|\G)p(\G|\A).
\end{equation}

We characterize this posterior using MH-MCMC.
The algorithm starts with an initial graph $\G$ which is sequentially updated through a proposal step followed by an accept reject step.
Our proposal is similar to the one used in \cite{Mukjerjee} and it changes
each edge $G_{ij}$ independently with probability 0 $< \tau <$ 1, or keeps it the same with probability 1 - $\tau$.
Note that higher values of $\tau$ makes the algorithm more exploratory. 
So, using, many values of $\tau$ and adjusting them, allows faster exploration of the space. Specifically, we pick edges at random and we or not change it. As mentioned, different $\tau$ leads to more noise added to the original data. The largest the $\tau$ the more edges are deleted and added. Using a variety of proposals based on $\tau$ value, can bring the acceptance rate to a certain level. 

\subsection{Making Decisions under Causal Graph Uncertainty}
\label{sec:dm_with_uncertainty_graphs}

The key to making decisions is estimating the expected value, $\cE{v(\Q)-c(\I)}{\doc{\I=\bi}}$, and the risk, $\cV{v(\Q)-c(\I)}{\doc{\I=\bi}}$, of an intervention $\doc{\I=\bi}$ as defined in \qref{eqn:causal}.
However, this program is problematic because of the potential unidentifiability and intractability issues.
To appreciate the complexity, let us expand the expected value.
We have:
$$
\cE{v(\Q)-c(\I)}{\doc{\I=\bi}} = \sum_\G \int \left(v(\bq) - c(\bi)\right)p(\bq|\doc{\I=\bi},\calD, \G)d\bq p(\G|\calD),
$$
where $p(\bq|\doc{\I=\bi},\calD, \G)$ is the causal effect obtained by applying Pearl's do-calculus \citep{Pearl} on the joint posterior predictive conditioned on the graph, see~\qref{eqn:predictive}, and $p(\G|\calD)$ is our posterior about the causal graph, which is characterized by MCMC sampling, see Sec.~\ref{sec:posterior_G}.
Unfortunately, this expression is nonsensical because the causal effect $p(\bq|\doc{\I=\bi},\calD, \G)$ is not always available.
Namely, the causal effect may either be unidentifiable from observational data or the expression provided by do-calculus for its calculation may be computationally intractable.


To overcome these difficulties we introduce the following definition.
Let $\G$ be a causal graph and consider an intervention-quantity of interest pair $(\I, \Q)$.
We say that the pair $(\I, \Q)$ is $\G$-good if the causal effect $p(\bq|\doc{\I=\bi},\calD, \G)$ is identifiable and there exists a computationally tractable formula, e.g., through the backdoor criterion, to calculate it.
Otherwise, we say that $(\I, \Q)$ is $\G$-bad, or just bad.

Note that one can easily calculate the expected value and the risk of a intervention conditioned $\G$ if the pair $(\I, \Q)$ is $\G$-good.
However, if $(\I, \Q)$ is $G$-bad, then the expected value and risk are unavailable.
In such a case the decision maker should model their belief about the value of an intervention with unknown outcomes and take this belief into account.
To this end, we define the random value random variable:
\begin{equation}
    V = \begin{cases}
    v(\Q),\text{if}\;(\I,\Q)\;\text{is}\;\G\text{-good}\\
    V_{\text{bad}},\;\text{otherwise},
    \end{cases}
\end{equation}
where $V_{\text{bad}}$ is a random variable modeling the value of an intervention with unknown consequences.
In our numerical examples we take $V_{\text{bad}}$ to be a Gaussian:
\begin{equation}
    V_{\text{bad}} \sim N(v_{\text{bad}}, r_{\text{bad}}),
\end{equation}
with $v_{\text{bad}}$ and $r_{\text{bad}}$ being decision-maker parameters corresponding to the expected value and variance of an intervention with unknown consequences.

We have now reached the final position of this paper: A rational decision maker should pick interventions that lie on the Pareto front of the following multi-objective optimization problem:
\begin{align}
    & \max_{\bi} \cE{V - c(\I)}{\doc{\I=\bi},\calD}, \\
    & \min_{\bi} \cV{V - c(\I)}{\doc{\I=\bi},\calD}.
\end{align}
In practice, we estimate both the expectation and the variance in the expressions using Bayesian Model Averaging \cite{hoeting1999bayesian} over graphs sampled from the posteriors (linear combination of the posteriors) using the MH-MCMC algorithm described above. 
Bayesian model averaging is what differentiates further from previous methodologies. Essentially, by considering all the possible causal relationships compatible with the prior beliefs and the data one gets a more robust and trustworthy methodology, meaning that either we approximate the true model if we have produce many networks and the true networks is among them of we approximate the closest model to the true model based on the available-candidate models, leveraging all the available information.




\section{Numerical Results}\label{Sim}
In this section, we present three simulation studies in this section. All these studies are purposefully free of spurious causal peer effects, see \cite{angrist2014perils,fafchamps2015causal,heterogenetwork,mendolia2018heterogeneous,tincani2018heterogeneous}. 
All variables are exogenous and instrumental variables have been used to sort out the problem of spurious effects.
We explore through the MH-MCMC algorithm all the posterior over causal structures. Then, we use the Bayesian Model Averaging method to resolve uncertainty and characterize the Pareto front of expected value and risk for each possible intervention.

\subsection{Example: Software Feature Selection \& Intervention}
Consider a team of data scientists at a software company that needs to identify causal relationships among data variables to decide which one is more important for programmers to modify and, therefore, increase software usage. Data scientists usually consider probabilistic relationships (e.g., regressions) for such analysis. Their decisions often fall short of expected impact, and programmers have to rely on their intuition to achieve better results. 
Our objective is to rank potential interventions according to their expected value and risk using only vague prior knowledge and observational data. 
To preserve the anonymity of the firm providing the data, we use an anonymized simulated dataset based on existing variable names.

\begin{figure}
  \centering
  \includegraphics[scale=0.8]{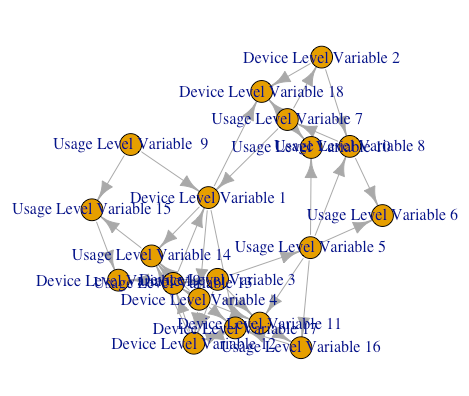}
  \caption{DAG from PC algorithm for all available transformed data.}
\end{figure}
\begin{figure}
  \includegraphics[scale=0.6]{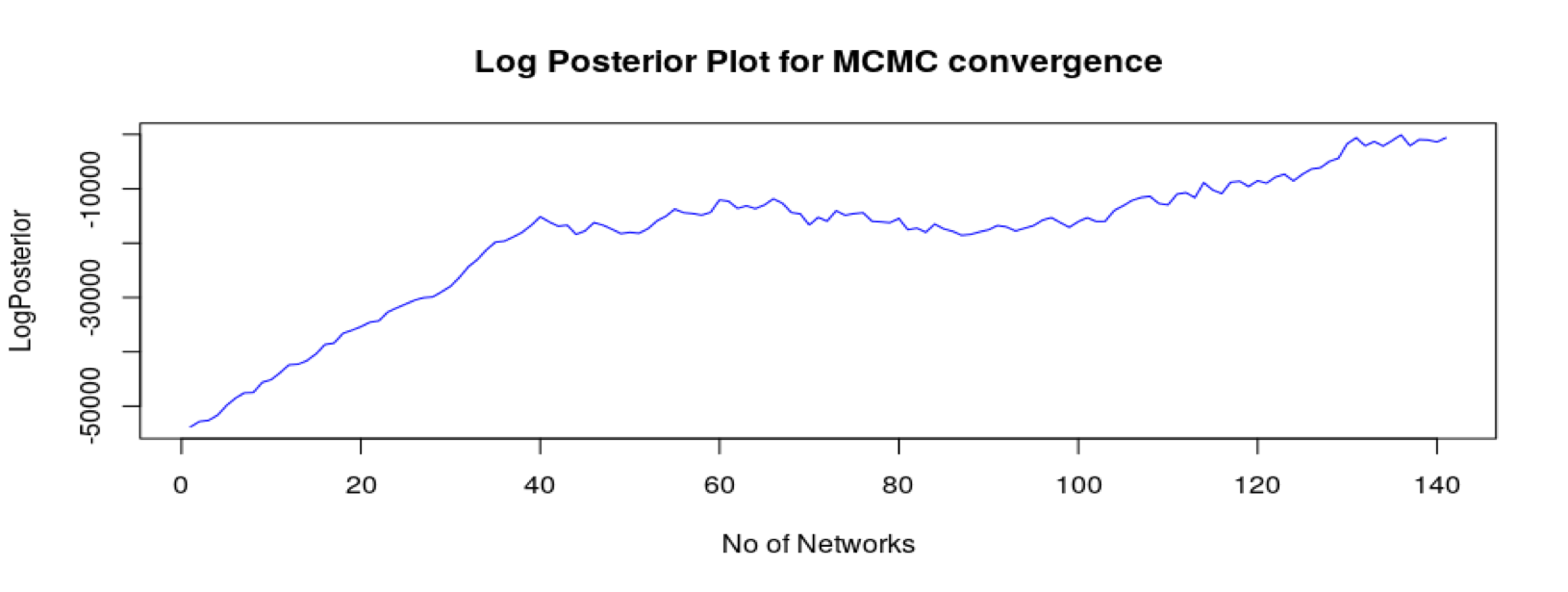}
  \caption{Plot which shows that MCMC converges. From all 141 samples (and 109 after the burn in period) Device Level Variable 16 exists in 59. We will show the results of $p(qoi \mid intervention var=i, data) = p(Usage Level Variable 1 \mid Device Level Variable 16 = i,data)$}
\end{figure}
\begin{figure}
  \centering
  \includegraphics[scale=0.4]{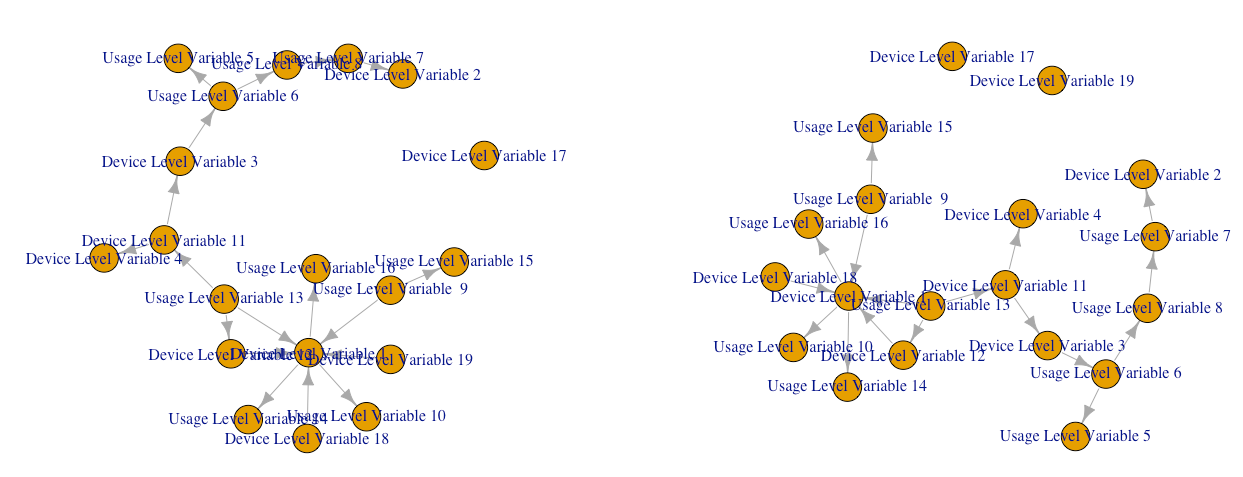}
  \caption{Two sampled DAGs consistent with data and prior belief, obtained through MH-MCMC process.}
\end{figure}
\begin{table}
\centering
\fbox{%
\begin{tabular}{| l  l  l l|}
\hline
Variables  & Name & Expected Value & Variance
 \\
\hline            
Qo1: V1 &  Usage Level Variable 1 & & \\
\hline            
IV1: V9 & Usage Level Variable 9 & 0.5549149 &0.048272\\
\hline            
IV2: V12& Device Level Variable 12 & 0.4138726 &0.0828718\\
\hline    
IV3: V13& Usage Level Variable 13 & 0.4008835 &0.03129\\
\hline    
IV4: V17& Device Level Variable 17 & 0.3937593 &0.0528716\\
\hline    
\end{tabular}}
\caption{MCMC Causal expectation values for the coefficients in the robust approach we present. For example, if we need to choose one intervention variable to tweak in order to affect QoI more, IV1 is more profitable choice than tweaking IV2, IV3 or IV4, when the same budget is available.}
\end{table}
\begin{table}
\centering
\fbox{%
\begin{tabular}{| l  l  l l |}
\hline
Variable & Potential Budget &  Expectation &  Variance  \\
\hline
Usage Level Variable 17 & 100000  & 17190 & 3483\\
\hline            
Usage Level Variable 13 & 100000 & 31627 & 3781\\
\hline            
Usage Level Variable 9 & 100000 & 54374  & 3282 \\
\hline            
Device Level variable 12 & 100000  &46872 & 3382  \\
\hline  
\end{tabular}
}
\caption{Causal Decision making for variable's potential budgets. Usage Level Variable 17 is the most expensive option to change in our case study, because they will remain almost 17000 dollars from our budget. Usage Level variable 9 is the least expensive option to change. The amount of information that will be changed is given by the causal coefficient. }
\end{table}
Clearly, in figure 2 we can see all causal relationships that arise using our algorithm. Running the described robust approach results in 141 potential causal graphs, as we can see from MH-MCMC log-Posterior convergence diagram (figure 3). Dataset consists of almost one hundred thousand rows of panel data with 19 attributes. Two of those attributes can be seen in figure 4. Table 2 presents some causal expected and variance values of the causal coefficients for the robust approach. Those results can be used to easily check the difference between applying a causal discovery algorithm against a probabilistic algorithm, such as Chow-Liu.

In table 2 we see how the final outcome 
is most notably caused by tweaking Usage Level Variable 9,
 Device Level Variable 12,
 Usage Level Variable 13,
and Usage Level Variable 17. 
For example, in order to increase the final outcome 
we can invest in increasing the access to a higher quality service or a more reliable device. Table 3 presents a case study for potential budgets that a company is willing to invest for each of those intervention variables based on the method presented in this paper, with expectation and variance for causal probabilities in order to make decisions.
\subsection{Application: Aviation Industry Scheduling}

In this simulation analysis, we use an airline data set to evaluate the models that consider long-term temporal trends for the airline scheduling process. From a raw data set, which contains 1.3 million direct flights
(data points) defined by forty different features over a one-year period provided by a major U.S. airline \citep{ogunsina2019hidden}, this section describes the methods used to abstract and encode different data features in the data set in order to achieve high fidelity
probabilistic graphical models. Since Bayesian network algorithms operate better with continuous data, all categorical values in the data set must be encoded into usable and valid continuous data before being used in a Bayesian inference.

Turn around is defined as the process of or time needed for loading, unloading, and servicing an aircraft. From our analysis, we see that by tweaking the total on board passengers, the scheduled aircraft type and the scheduled aircraft push back time at origin we can change with high probability the variables  scheduled aircraft push back time at origin,  the scheduled turnaround period and the flag to indicate scheduled route originator, accordingly. All the minor and major probabilistic and causal relationships are illustrated in figure 5 and presented in tables 5 and 6, respectively. Table 6 presents a case study for potential budgets that a company is willing to invest for each of those intervention variables for a specific quantity of interest.

\begin{figure}[ht]
\includegraphics[scale=0.7]{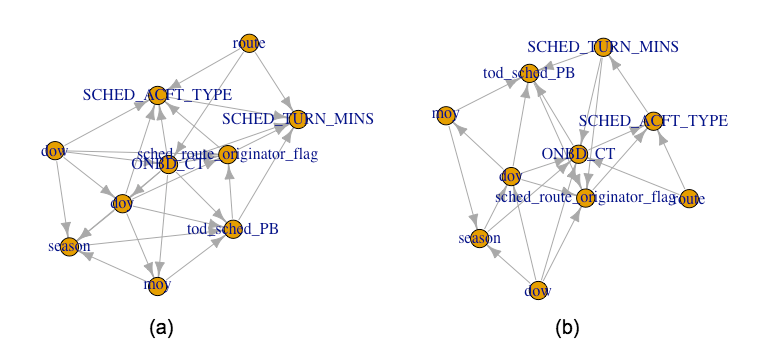}
\caption{a) Random-restart Hill-Climbing algorithm for TAS probabilistic relationships. b) TAS causal relationships from PC algorithm. Here we use the robust approach to get all the observed sampled graphs.}
\end{figure}

\begin{table}
\centering
\fbox{%
\begin{tabular}{| l  l |}
\hline
Dependent Variable  & Independent Variable \\
\hline            
tod\_sched\_PB & season \\
\hline
tod\_sched\_PB  & ONBD\_CT  \\
\hline
SCHED\_TURN\_MINS & ONBD\_CT   \\
\hline     
SCHED\_TURN\_MINS & sched\_rout\_originator\_flag   \\
\hline  
SCHED\_TURN\_MINS & SHED\_ACFT\_TYPE   \\
\hline  
SCHED\_ACFT\_TYPE & ONBD\_CT   \\
\hline  
SCHED\_TURN\_MINS & sched\_rout\_originator\_flag   \\
\hline  
sched\_rout\_originator\_flag  & ONBD\_CT  \\
\hline  
sched\_rout\_originator\_flag  & tod\_sched\_PB   \\
\hline  
\end{tabular}}
\caption{Probabilistic relationships in TAS}
\end{table}

\begin{table}
\centering
\fbox{%
\begin{tabular}{| l  l  l l|}
\hline
Quantity of Interest  & Intervention Variable & Expected Value & Variance \\
\hline
tod\_sched\_PB  & ONBD\_CT & 0.72652 & 0.09452 \\
\hline  
SCHED\_TURN\_MINS & SHED\_ACFT\_TYPE  & 0.70991 & 0.08271\\
\hline  
SCHED\_ACFT\_TYPE & route  & 0.526517 & 0.0629815  \\
\hline  
SCHED\_ACFT\_TYPE & sched\_rout\_originator\_flag  &0.407362 & 0.0762865 \\
\hline  
SCHED\_ACFT\_TYPE &  ONBD\_CT  & 0.572635 & 0.0842973 \\
\hline  
sched\_rout\_originator\_flag  & SCHED\_TURN\_MINS & 0.82787& 0.09827675 \\
\hline  
sched\_rout\_originator\_flag  & tod\_sched\_PB  & 0.910312 & 0.128271 \\
\hline  
\end{tabular}}
\caption{ Causal expectation and variance values for the coefficients in TAS.}
\end{table}

\begin{table}
\centering
\fbox{%
\begin{tabular}{| l  l  l l |}
\hline
Variable & Potential Budget & Expection & Variance \\ \hline
$sched\_route\_originator\_flag$& 100000 &  42716 & 2671 \\
\hline            
$ONBD\_CT$ & 100000 & 30434 & 3827 \\
\hline            
\end{tabular}}
\caption{Causal Decision making for variable's potential budgets. $sched\_route\_originator\_flag$ is less expensive option to change than $ONBD\_CT$ for the variable $SCHED\_ACFT\_TYPE$. }
\end{table}

\subsection{Ground Truth Comparison: The LUCAS Dataset}

The objective of this experiment is to verify how our approach is compared with a real ground truth example. So we consider an $\mathcal{M}$-closed case and we regard as ground truth an experiment for which we already know the results, in our case the causal dependencies. Thus, we are able to validate our model. We use LUCAS (LUng CAncer Simple set) for this, which contains simulated data created by causal Bayesian networks with binary variables. \cite{daza2020lucas} mentions that these examples are entirely made up and are mainly used for illustrating purposes.

\begin{figure}
\center
  \includegraphics[scale=0.55]{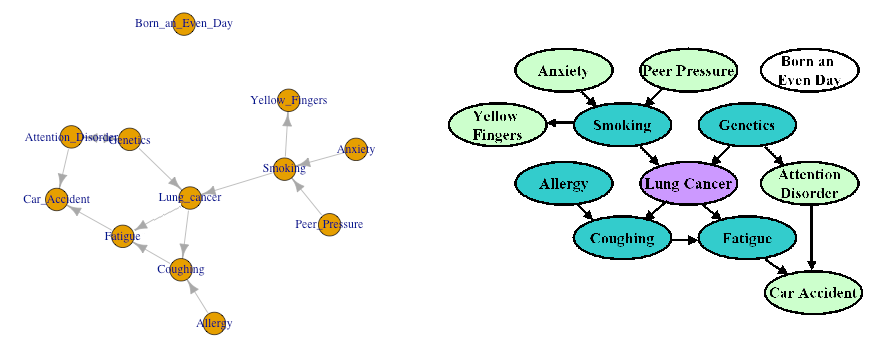}
  \caption{PC algorithm compared with the a ground Truth example (LUCAS).}
\end{figure}

\begin{figure}
\center
  \includegraphics[scale=0.45]{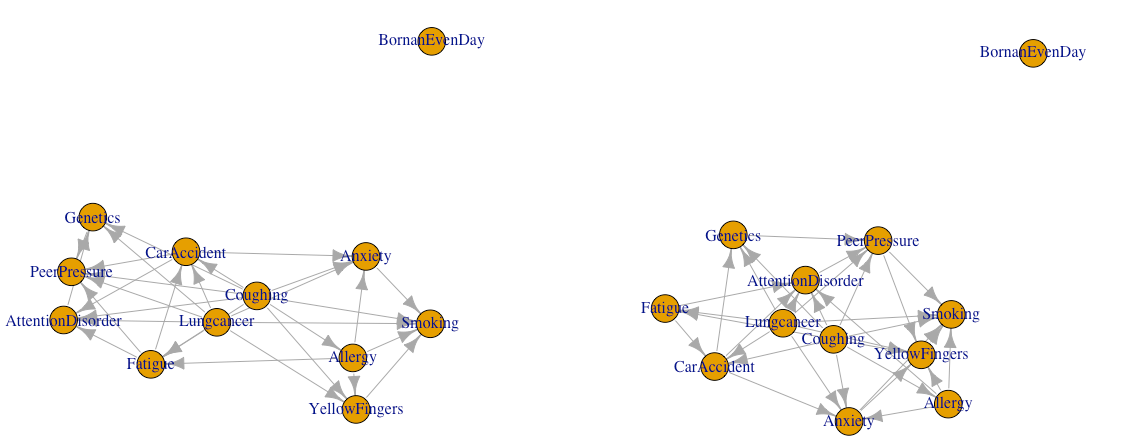}
  \caption{Two sample networks that were produced from the 54 networks produced.}
\end{figure}

\begin{table}
\centering
\fbox{%
\begin{tabular}{| l c  c c|}
\hline
Variables   & Ground Truth probability & Expected Value & Variance\\
\hline    
QoI1: Lung cancer=T &  &  &  \\
\hline    
IV1: Genetics=F, Smoking=F & 0.23146 & 0.21328 & 0.038172 \\
\hline    
IV2: Genetics=T, Smoking=F & 0.86996 & 0.872615 & 0.035917 \\
\hline    
IV3: Genetics=F, Smoking=T & 0.83934 & 0.083217& 0.009818 \\
\hline    
IV4: Genetics=T, Smoking=T & 0.99351 & 0.99382 & 0.006281\\
\hline            
QoI2: Coughing=T &  &  &   \\
\hline
IV1: Allergy=T, Lung cancer=F & 0.64592 & 0.63827 & 0.05473\\
\hline
IV2: Allergy=F, Lung cancer=T & 0.7664 & 0.7695 & 0.02833\\
\hline
IV3: Allergy=T, Lung cancer=T & 0.99947 & 0.9995 & 0.0003\\
\hline
QoI3: Fatigue=T & & &\\
\hline
IV1: Lung cancer=F, Coughing=F & 0.35212 & 0.33648 &0.08827\\
\hline
IV2: Lung cancer=T, Coughing=F & 0.56514 & 0.57362& 0.07726\\
\hline
IV3: Lung cancer=F, Coughing=T & 0.80016 & 0.77372 & 0.079827\\
\hline
IV4: Lung cancer=T, Coughing=T &0.89589 & 0.91824 & 0.06899\\
\hline
QoI4: Smoking=T& & & \\
\hline
IV1: Peer Pressure=F, Anxiety=F & 0.43118 & 0.44838 & 0.030281\\
\hline
IV2: Peer Pressure=T, Anxiety=F & 0.74591 & 0.74285 & 0.02828 \\
\hline
IV3: Peer Pressure=F, Anxiety=T & 0.8686 & 0.8693  & 0.01128 \\
\hline
IV4: Peer Pressure=T, Anxiety=T & 0.91576 & 0.91482 & 0.01928 \\
\hline
\end{tabular}}
\caption{MH-MCMC Causal expectation and variance values for the coefficients for robust approach vs the ground truth approach for LUCAS data.}
\end{table}

For the regimes displayed in Table 7, we examine the variability of our approach by running it 11000 times and we compute the mean and the variance compared with the ground truth. Specifically, 54, 53, 55 and 53 sampled causal directed acyclic network instances were produced for QoI1, QoI2, QoI3 and QoI4 accordingly. We regard as ground truth the output from a the LUCAS experiment described above. As expected, table 7 confirms the reliability of our approaches since the expected values and the variances computed of our exhausting method fall into the set of values where the results from the ground truth are concentrated. The estimated variances can be regarded as very small, which happens because of the nature of the data. By, comparing the two approaches, finding causal dependencies and our approach, we can see the following:

\begin{itemize}
    \item Our approach approximate the ground truth result (since we have, intentionally, an $\mathcal{M}$-closed problem a large enough sample of networks, produced from the MH-MCMC) and the examiner is able to check the variability of those results, as well. 
    \item There are some sampled networks, as in figure 7, where the data does not always produce the right causal dependencies directions. For instance, in QoI3, Fatigue and Lung Cancer might have reverse causal effect. This is happening both because of the nature of the data and way they are fitted in the variable selection criterion. Those association, which are controversial could not be capture easily from those mechanisms and our method provides a further insight to the practitioner. Thus, it can influence, in a helpful way, the subjective prior knowledge of the examiner in order to avoid those situations. 
\end{itemize}

\subsection{Practitioner Prior Analysis}

Bayesian inference is drawn from the posterior distribution. However, these values are heavily affected by the choice of the prior, even if you have decided to go for an uninformed one. In order the practitioner to convince the audience that your choice of the prior does or does not lead your results a priori in a way that would diminish the value of your results and showing that your results were or were not heavily influenced by your choice of prior is simply to show that your results hold or not when changing the prior. Thus, this is happening because, in many situations, the analyst may not want this term to contribute
to model evaluation. Furthermore, there is a tension between the need to define undefined priors in order to protect their power and the fact that diffuse priors will result in arbitrary large and negative model results for real-valued parameters.

Here, we provide a simple example motivated by the app company's data in 4.1. We change the practitioner's prior belief of the variable Usage\_Level\_Variable\_8 against Usage\_Level\_Variable\_7. Figure 8 indicates how the prior due to subjective knowledge influences (Usage\_Level\_Variable\_8) the causal coefficient value of some other variable of interest (Usage\_Level\_Variable\_7).
\begin{figure}
  \includegraphics[scale=0.6]{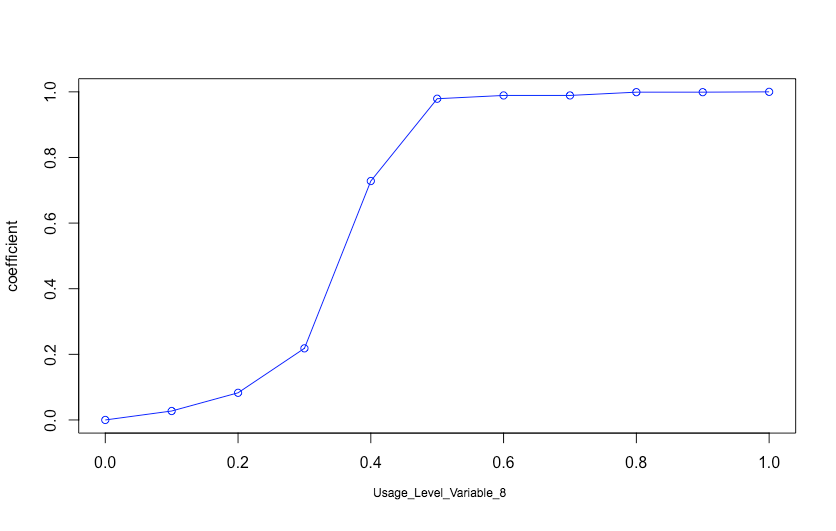}
  \caption{Prior sensitivity analysis of Usage\_Level\_Variable\_8 vs the standardized causal coefficient that influences Usage\_Level\_Variable\_7}
\end{figure}
The purpose of this design is to provide the examiner with a tool to criticize/visualize with more transparency his decisions vs reality. In order to turn the results above to a decision the practitioner has to resort to subsections that refer to causal decision making settings. From figure 6 we see very easily that if the practitioner does not believe a priori that Usage\_Level\_Variable\_8 has an impact on Usage\_Level\_Variable\_7 is wrong and very likely due to the data to take the wrong decision. That happens because the causal coefficient when we are based only on the data (prior is 0.5) is close to 0.97. Moreover, we can easily check that if he is over-optimistic and believes that changes Usage\_Level\_Variable\_8 will impact Usage\_Level\_Variable\_7 does not different much from the information given by the data. Values of the causal coefficient due to a prior larger than 0.5 are very close.

\section{Discussion}

We have presented a method which takes advantage of the network structure and constructs the most resilient joint distribution using Bayesian Model Averaging for $\mathcal{M}$-closed, complete and open problems. This distribution leverages all available causal dependencies between variables, from all candidate networks produced by an MH-MCMC, a practitioner can be provided with. A loose interpretation of our approach, could lead to a causal variable selection criterion. Computationally, the likelihood that comes from the data does not dominate the  binomial prior due to the relative likelihood. In contract with regression models models, Bayesian networks when they are dealing with network data are not that flexible and can not capture subtle interrelationships. Bayesian networks are not misleading, disclosing the structure and the probabilistic relationships between the variables.

In this paper, we extended out approach to causal inference in order to take advantage the most information for decision making problems. We address and formulate the problem of how can we take advantage of those important causal dependencies in order to make more resilient decisions. We argue that after
we average out all the network structure and the coefficients,
we will end up with the true or most closely to the true joint distribution that the will make the practitioner understand and describe summary statistics of the network data causal dependencies better. So, given all prior vague information and constructing the network that
characterizes their probabilistic relationships, we end up using M-H MCMC to provide all candidate causal models which after Bayesian Model Averaging approximate the model
which is more likely to survive, following the rationale that the most important models survive the most after
averaging out all the uncertainty that is encoded by the structure and
the regression coefficients.

Specifically, this paper proposes a methodology for identifying the most of information based on causal networks. We know that probabilistic and causality notion might sometimes overlap but are different. Taking into account both notions combined with vague prior beliefs of the practitioners leads to better results in decision making problems. Our simulation studies lead to several conclusions. The basic one is that the approach is cautious and trustworthy. Given the reliability of these estimates we are optimistic about the results, and are currently experimenting with this approach. Finally, these results indicate that the approach can be a reliable method for detecting causal dependencies for solving decision making problems.  

As alluded to in the paper, the core ideas presented
here can be applied more broadly since the complexity of the problem is very high. This is one direction for future work we are currently on. Resorting to scalable solutions \cite{Gao} is one options which could speed up a lot our method and apply it both as model selection in the whole space. Moreover, another limitations of our method, which can be examined are constructing methods for identifying latent variables and spurious causal relationships of the data and embed them to our approach.

\bibliographystyle{chicago}
\bibliography{mybib}

\begin{thebibliography}{}

\bibitem[\protect\citeauthoryear{Angrist}{Angrist}{2014}]{angrist2014perils}
Angrist, J.~D. (2014).
\newblock The perils of peer effects.
\newblock {\em Labour Economics\/}~{\em 30}, 98--108.

\bibitem[\protect\citeauthoryear{Athey and Wager}{Athey and
  Wager}{2019}]{athey2019estimating}
Athey, S. and S.~Wager (2019).
\newblock Estimating treatment effects with causal forests: An application.
\newblock arXiv preprint arXiv:1902.07409.

\bibitem[\protect\citeauthoryear{Azoury}{Azoury}{1985}]{azoury1985bayes}
Azoury, K.~S. (1985).
\newblock Bayes solution to dynamic inventory models under unknown demand
  distribution.
\newblock {\em Management science\/}~{\em 31\/}(9), 1150--1160.

\bibitem[\protect\citeauthoryear{Banerjee, Breza, Chandrasekhar, Duflo,
  Jackson, and Kinnan}{Banerjee et~al.}{2021}]{banerjee2021changes}
Banerjee, A., E.~Breza, A.~G. Chandrasekhar, E.~Duflo, M.~O. Jackson, and
  C.~Kinnan (2021).
\newblock Changes in social network structure in response to exposure to formal
  credit markets.
\newblock Technical report, National Bureau of Economic Research.

\bibitem[\protect\citeauthoryear{Bapna, Ramaprasad, Shmueli, and Umyarov}{Bapna
  et~al.}{2016}]{bapna2016one}
Bapna, R., J.~Ramaprasad, G.~Shmueli, and A.~Umyarov (2016).
\newblock One-way mirrors in online dating: A randomized field experiment.
\newblock {\em Management Science\/}~{\em 62\/}(11), 3100--3122.

\bibitem[\protect\citeauthoryear{Berger}{Berger}{2013}]{berger2013statistical}
Berger, J.~O. (2013).
\newblock {\em Statistical decision theory and Bayesian analysis}.
\newblock Springer Science \& Business Media.

\bibitem[\protect\citeauthoryear{Chu, Shanthikumar, and Shen}{Chu
  et~al.}{2008}]{chu2008solving}
Chu, L.~Y., J.~G. Shanthikumar, and Z.-J.~M. Shen (2008).
\newblock Solving operational statistics via a bayesian analysis.
\newblock {\em Operations Research Letters\/}~{\em 36\/}(1), 110--116.

\bibitem[\protect\citeauthoryear{Claeskens, Hjort, et~al.}{Claeskens
  et~al.}{2008}]{claeskens2008model}
Claeskens, G., N.~L. Hjort, et~al. (2008).
\newblock {\em Model selection and model averaging}.
\newblock Cambridge University Press.

\bibitem[\protect\citeauthoryear{Daza, Castillo, Escobar, Valencia, Pinz{\'o}n,
  and Arbel{\'a}ez}{Daza et~al.}{2020}]{daza2020lucas}
Daza, L., A.~Castillo, M.~Escobar, S.~Valencia, B.~Pinz{\'o}n, and
  P.~Arbel{\'a}ez (2020).
\newblock Lucas: Lung cancer screening with multimodal biomarkers.
\newblock In {\em Multimodal Learning for Clinical Decision Support and
  Clinical Image-Based Procedures}, pp.\  115--124. Springer.

\bibitem[\protect\citeauthoryear{Ertefaie, Asgharian, and Stephens}{Ertefaie
  et~al.}{2018}]{ertefaie2018variable}
Ertefaie, A., M.~Asgharian, and D.~A. Stephens (2018).
\newblock Variable selection in causal inference using a simultaneous
  penalization method.
\newblock {\em Journal of Causal Inference\/}~{\em 6\/}(1), 1--16.

\bibitem[\protect\citeauthoryear{Fafchamps}{Fafchamps}{2015}]{fafchamps2015causal}
Fafchamps, M. (2015).
\newblock Causal effects in social networks.
\newblock {\em Revue {\'e}conomique\/}~{\em 66\/}(4), 657--686.

\bibitem[\protect\citeauthoryear{Feng and Shanthikumar}{Feng and
  Shanthikumar}{2018}]{feng2018research}
Feng, Q. and J.~G. Shanthikumar (2018).
\newblock How research in production and operations management may evolve in
  the era of big data.
\newblock {\em Production and Operations Management\/}~{\em 27\/}(9),
  1670--1684.

\bibitem[\protect\citeauthoryear{Fragoso, Bertoli, and Louzada}{Fragoso
  et~al.}{2018}]{fragoso2018bayesian}
Fragoso, T.~M., W.~Bertoli, and F.~Louzada (2018).
\newblock Bayesian model averaging: A systematic review and conceptual
  classification.
\newblock {\em International Statistical Review\/}~{\em 86\/}(1), 1--28.

\bibitem[\protect\citeauthoryear{Friedman, Linial, Nachman, and Pe'er}{Friedman
  et~al.}{2000}]{Friedman}
Friedman, N., M.~Linial, I.~Nachman, and D.~Pe'er (2000).
\newblock Using bayesian networks to analyze expression data.
\newblock {\em Comput. Biol.\/}~{\em 7}, 601--620.

\bibitem[\protect\citeauthoryear{Gao and Wei}{Gao and Wei}{2018}]{Gao}
Gao, T. and D.~Wei (2018).
\newblock Parallel bayesian network structure learning.
\newblock In {\em International Conference on Machine Learning}, pp.\
  1685--1694.

\bibitem[\protect\citeauthoryear{Geiger and Heckerman}{Geiger and
  Heckerman}{1994}]{geiger1994learning}
Geiger, D. and D.~Heckerman (1994).
\newblock Learning gaussian networks.
\newblock In {\em Uncertainty Proceedings 1994}, pp.\  235--243. Elsevier.

\bibitem[\protect\citeauthoryear{Gibbard and Harper}{Gibbard and
  Harper}{1978}]{gibbard1978counterfactuals}
Gibbard, A. and W.~L. Harper (1978).
\newblock Counterfactuals and two kinds of expected utility.
\newblock In {\em Ifs}, pp.\  153--190. Springer.

\bibitem[\protect\citeauthoryear{Heckerman and Geiger}{Heckerman and
  Geiger}{2013}]{heckerman2013learning}
Heckerman, D. and D.~Geiger (2013).
\newblock Learning bayesian networks: a unification for discrete and gaussian
  domains.
\newblock arXiv preprint arXiv:1302.4957.

\bibitem[\protect\citeauthoryear{Heckerman, Geiger, and Chickering}{Heckerman
  et~al.}{1995}]{heckerman1995learning}
Heckerman, D., D.~Geiger, and D.~M. Chickering (1995).
\newblock Learning bayesian networks: The combination of knowledge and
  statistical data.
\newblock {\em Machine learning\/}~{\em 20\/}(3), 197--243.

\bibitem[\protect\citeauthoryear{Heckerman~D. and Chickering}{Heckerman~D. and
  Chickering}{2015}]{Heckerman}
Heckerman~D., D.~G. and D.~M. Chickering (2015).
\newblock Learning bayesian networks: the combination of knowledg eand
  statistical data.
\newblock {\em Machine Learning\/}~{\em 20}, 197--243.

\bibitem[\protect\citeauthoryear{Hoeting, Madigan, Raftery, and
  Volinsky}{Hoeting et~al.}{1999}]{hoeting1999bayesian}
Hoeting, J.~A., D.~Madigan, A.~E. Raftery, and C.~T. Volinsky (1999).
\newblock Bayesian model averaging: a tutorial.
\newblock {\em Statistical science\/}~{\em 14\/}(4), 382--401.

\bibitem[\protect\citeauthoryear{Huang, Sun, Chen, and Golden}{Huang
  et~al.}{2019}]{huang2019word}
Huang, N., T.~Sun, P.~Chen, and J.~M. Golden (2019).
\newblock Word-of-mouth system implementation and customer conversion: A
  randomized field experiment.
\newblock {\em Information Systems Research\/}~{\em 30\/}(3), 805--818.

\bibitem[\protect\citeauthoryear{Jung, Bapna, Ramaprasad, and Umyarov}{Jung
  et~al.}{2019}]{jung2019love}
Jung, J., R.~Bapna, J.~Ramaprasad, and A.~Umyarov (2019).
\newblock Love unshackled: Identifying the effect of mobile app adoption in
  online dating.
\newblock {\em MIS Quarterly\/}~{\em 43}, 47--72.

\bibitem[\protect\citeauthoryear{Kohavi, Deng, Frasca, Walker, Xu, and
  Pohlmann}{Kohavi et~al.}{2013}]{kohavi2013online}
Kohavi, R., A.~Deng, B.~Frasca, T.~Walker, Y.~Xu, and N.~Pohlmann (2013).
\newblock Online controlled experiments at large scale.
\newblock In {\em Proceedings of the 19th ACM SIGKDD international conference
  on Knowledge discovery and data mining}, pp.\  1168--1176.

\bibitem[\protect\citeauthoryear{Kohavi and Thomke}{Kohavi and
  Thomke}{2017}]{kohavi2017surprising}
Kohavi, R. and S.~Thomke (2017).
\newblock The surprising power of online experiments.
\newblock {\em Harvard Business Review\/}~{\em 95\/}(5), 74--82.

\bibitem[\protect\citeauthoryear{Koller and Friedman}{Koller and
  Friedman}{2009}]{Koller}
Koller, D. and N.~Friedman (2009).
\newblock {\em Probabilistic graphical models: principles and techniques}.
\newblock MIT press.

\bibitem[\protect\citeauthoryear{Livia~Shkozaa}{Livia~Shkozaa}{2019}]{heterogenetwork}
Livia~Shkozaa, D.~U. (2019).
\newblock Heterogeneity in network peer effects.
\newblock University of Munich.

\bibitem[\protect\citeauthoryear{Liyanage and Shanthikumar}{Liyanage and
  Shanthikumar}{2005}]{liyanage2005practical}
Liyanage, L.~H. and J.~G. Shanthikumar (2005).
\newblock A practical inventory control policy using operational statistics.
\newblock {\em Operations Research Letters\/}~{\em 33\/}(4), 341--348.

\bibitem[\protect\citeauthoryear{Mendolia, Paloyo, and Walker}{Mendolia
  et~al.}{2018}]{mendolia2018heterogeneous}
Mendolia, S., A.~R. Paloyo, and I.~Walker (2018).
\newblock Heterogeneous effects of high school peers on educational outcomes.
\newblock {\em Oxford Economic Papers\/}~{\em 70\/}(3), 613--634.

\bibitem[\protect\citeauthoryear{Mukherjee and Speed}{Mukherjee and
  Speed}{2008}]{Mukjerjee}
Mukherjee, S. and T.~P. Speed (2008).
\newblock Network inference using informative priors.
\newblock {\em Proceedings of the National Academy of Sciences\/}~{\em
  105\/}(38), 14313--14318.

\bibitem[\protect\citeauthoryear{Murphy, Mian, et~al.}{Murphy
  et~al.}{1999}]{Murphy}
Murphy, K., S.~Mian, et~al. (1999).
\newblock Modelling gene expression data using dynamic bayesian networks.
\newblock Technical report, Technical report, Computer Science Division,
  University of California.

\bibitem[\protect\citeauthoryear{Ogunsina, Papamichalis, Bilionis, and
  DeLaurentis}{Ogunsina et~al.}{2019}]{ogunsina2019hidden}
Ogunsina, K.~E., M.~Papamichalis, I.~Bilionis, and D.~A. DeLaurentis (2019).
\newblock Hidden markov models for pattern learning and recognition in a
  data-driven model for airline disruption management.
\newblock In {\em AIAA Aviation 2019 Forum}, pp.\  3508.

\bibitem[\protect\citeauthoryear{Pearl}{Pearl}{2009}]{Pearl}
Pearl, J. (2009).
\newblock {\em Causality}.
\newblock Cambridge university press.

\bibitem[\protect\citeauthoryear{Ramamurthy, George~Shanthikumar, and
  Shen}{Ramamurthy et~al.}{2012}]{ramamurthy2012inventory}
Ramamurthy, V., J.~George~Shanthikumar, and Z.-J.~M. Shen (2012).
\newblock Inventory policy with parametric demand: Operational statistics,
  linear correction, and regression.
\newblock {\em Production and Operations Management\/}~{\em 21\/}(2), 291--308.

\bibitem[\protect\citeauthoryear{Robert}{Robert}{2007}]{robert2007bayesian}
Robert, C. (2007).
\newblock {\em The Bayesian choice: from decision-theoretic foundations to
  computational implementation}.
\newblock Springer Science \& Business Media.

\bibitem[\protect\citeauthoryear{Rosenbaum and Rubin}{Rosenbaum and
  Rubin}{1983}]{rosenbaum1983central}
Rosenbaum, P.~R. and D.~B. Rubin (1983).
\newblock The central role of the propensity score in observational studies for
  causal effects.
\newblock {\em Biometrika\/}~{\em 70\/}(1), 41--55.

\bibitem[\protect\citeauthoryear{Smith, Yu, Smulders, Hartemink, and
  Jarvis}{Smith et~al.}{2006}]{Smith}
Smith, V.~A., J.~Yu, T.~V. Smulders, A.~J. Hartemink, and E.~D. Jarvis (2006).
\newblock Computational inference of neural information flow networks.
\newblock {\em PLoS computational biology\/}~{\em 2\/}(11), e161.

\bibitem[\protect\citeauthoryear{Spirtes, Glymour, Scheines, Kauffman, Aimale,
  and Wimberly}{Spirtes et~al.}{2000}]{Spirtes}
Spirtes, P., C.~Glymour, R.~Scheines, S.~Kauffman, V.~Aimale, and F.~Wimberly
  (2000).
\newblock Constructing bayesian network models of gene expression networks from
  microarray data.
\newblock Carnegie Mellon University.

\bibitem[\protect\citeauthoryear{Sun, Gao, and Jin}{Sun
  et~al.}{2019}]{sun2019mobile}
Sun, T., G.~Gao, and G.~Z. Jin (2019).
\newblock Mobile messaging for offline group formation in prosocial activities:
  A large field experiment.
\newblock {\em Management Science\/}~{\em 65\/}(6), 2717--2736.

\bibitem[\protect\citeauthoryear{Sun, Shi, Viswanathan, and Zheleva}{Sun
  et~al.}{2019}]{sun2019motivating}
Sun, T., L.~Shi, S.~Viswanathan, and E.~Zheleva (2019).
\newblock Motivating effective mobile app adoptions: Evidence from a
  large-scale randomized field experiment.
\newblock {\em Information Systems Research\/}~{\em 30\/}(2), 523--539.

\bibitem[\protect\citeauthoryear{Sun, Viswanathan, and Zheleva}{Sun
  et~al.}{2021}]{tianshu21}
Sun, T., S.~Viswanathan, and E.~Zheleva (2021).
\newblock “creating social contagion through firm mediated message design:
  Evidence from a randomized field experiment”.
\newblock {\em Management Science\/}~{\em 67\/}(2), 808--827.

\bibitem[\protect\citeauthoryear{Tincani}{Tincani}{2018}]{tincani2018heterogeneous}
Tincani, M.~M. (2018).
\newblock Heterogeneous peer effects in the classroom.
\newblock Technical report, Working Paper.

\bibitem[\protect\citeauthoryear{Vansteelandt, Bekaert, and
  Claeskens}{Vansteelandt et~al.}{2012}]{vansteelandt2012model}
Vansteelandt, S., M.~Bekaert, and G.~Claeskens (2012).
\newblock On model selection and model misspecification in causal inference.
\newblock {\em Statistical methods in medical research\/}~{\em 21\/}(1), 7--30.

\bibitem[\protect\citeauthoryear{Yao, Vehtari, Simpson, Gelman, et~al.}{Yao
  et~al.}{2018}]{yao2018using}
Yao, Y., A.~Vehtari, D.~Simpson, A.~Gelman, et~al. (2018).
\newblock Using stacking to average bayesian predictive distributions (with
  discussion).
\newblock {\em Bayesian Analysis\/}~{\em 13\/}(3), 917--1007.

\end{thebibliography}

\section*{Online Appendix}

\subsection*{Causality in experimental vs observational designs}

Rubin's framework, or potential outcome framework, is about experimental data, where the treatment is assigned randomly by the practitioners to the units. An example is randomized controlled trials for finding the effects of a vaccine or a medication or more specifically in microfinance consider the experiment of \cite{banerjee2021changes} in Hyderabad, where
institution randomly selected neighborhoods to enter first. Collecting data for those experiments is time-consuming and requires to construct the experiment in the right way (e.g. following unconfoundess hypothesis for units), in order we can avoid issues that arise e.g. through the selection of the estimand, like heterogeneity in \cite{athey2019estimating}. Observational studies concern cases where the treatment is not assigned randomly by the practitioner. One example is collecting data from past data like social media. The key issue with retrospective research is that the procedure or exposure is not randomized, and the estimation of the impact of the exposure on an outcome can be skewed by confounders, which are factors that are involved in both the exposure and the outcome. Propensity ratings, such as those used in \cite{rosenbaum1983central}, are widely being used to control for confounding by estimating the chance of having the observed exposure amount provided the covariates. There are more methods that can use permutation test for observatonal studies like Difference in Differences. Here, we follow Pearl's framework. The main difference is that through DAG's and methods like backdoor criterion he tries to control the confounders conditioning on a network structure (e.g. blocking relationships could be achieved). Mathematically Pearl framework considers distributions (functions) that are related with conditional dependences and independences e.g. what about the conditional independences of the output? What are the parameters of the potential outcomes (in his terminology counterfactuals). He does not care about data, only about distributions. The basic advantage of experimental  observational studies is that it is dangerous for the latter to used in cases such as testing a new drug, or clinical trials. On the other hand, the main advantage of observational studies is that they are not time-consuming and can be used in cases where random assignment is unethical. We need to hightlight that, in both cases, unobserved confounders could be an issue. 

Intuitively, Pearl is concerned with the question that we have some distributions, we want to see-check the causal relationships what lead us here. His reasoning is based on a hypothesis that the data comes from a hypothetical causal graph, built from the data of the distributions. The process of learning causal consequences where the therapy is not under our influence and instead happens in a "normal" or "observational" regime. The aim is to link how evidence from the observational regime can be used to infer results from the interventional regime. In short, the framework is that of causal graphical models \cite{Pearl,Spirtes}, which uses directed acyclic graphs (DAGs) to represent causal dependencies. The framework is popular in computer science in estimating a causal structure. 

Discovering the causal model from the data, defining the causal effect when the causal structure is identified, and predicting an observable causal effect from the data are the three sub-areas of causal inference.  For the first case approximating algorithms are used (e.g. pc algorithm). For the second case we know how to calculate if a causal effect is identifiable or not (dc algorithm). Though, estimating automatically an identifiable causal effect is an active topic of research. In contrast with backdoor and frontdoor criteria, practitioners have to proceed by themselves.

Under the assumption that the data follows a causal graph (specifying an underlying structural equation model), there are situations where you can compute $P(Y\mid do(X=x))$ solely based on observed data:

\begin{itemize}
\item If $S$ is a subset of the variables that “blocks” all the causal paths from $Y$ to $X$ (backdoor paths) then you can write (backdoor theorem): $P(Y\mid do(X=x))=\sum_s P(Y\mid S=s,X=x)P(S=s)$.
\item If $M$ is a subset of the variables that “blocks” all the causal paths from $X$ to Y(frontdoor paths) then you can write (frontdoor theorem): $P(Y\mid do(X=x))=\sum_m P(Y=y \mid do(M=m)) P(m \mid do(X=x))$.
\item If $X$ blocks all the causal paths from $S$ to $Y$, then you can use $S$ as an instrumental variable to compute $E(Y\mid do(X=x))$ under some heavy linearity assumptions.
\end{itemize}

The causal graphical models framework which is used here, uses directed acyclic graphs (DAGs) to
represent causal dependencies. This framework is popular in computer science, however, it is not well suited
to our problem since we do not aim at estimating a
causal structure but rather the “marginal” peer influence causal effects through a randomized experiment.

The back-door and front-door conditions, as well as instrumental variables, are both adequate for estimating causal effects from probabilistic distributions. However they are not required, and appropriate and sufficient conditions for the identifiability of causal effects are technically feasible but lack a pleasant glib type.

\subsection*{PC algorithm}

One of the earliest methods employed for structure learning in probabilistic graphical methods is called the PC Algorithm  \citep{Spirtes}. Following is an intuitive explanation of this constraint-based structure learning method \citep{Koller}. 

First, an approximate undirected version of the directed acyclic graphical (DAG) structure underlying the data is estimated. The algorithm begins the approximation procedure assuming a complete undirected graph. Subsequently, each tuple of three vertices is checked for conditional independence. For instance, for vertices $v_{1}, v_{2}, v_{3}$, one of the conditional independence checks would be $v_{1} \indep v_{3} ~|~ v_{2}$. If such a vertex/set of vertices is found, then the edge between $v_{1} \& v_{3}$ is deleted. 

Second, once the approximate graph is ascertained, \textit{common effect} directionality for each tuple of three vertices where number of edges is two, is determined. Essentially, for each such tuple, there will always exist a vertex in the middle, having two edges connected to it. If such a vertex is not in the \textit{separation} set\footnote{Separation sets are nothing but sets of vertices that enforce conditional independence on other two vertices connected to it.} of the other two vertices, then the direction of edges is such that the edges flow into the vertex in the middle, thus signifying common effect (e.g., $v_{1} \rightarrow v_{2} \leftarrow v_{3}$). 

Third, using the partially directed graph from the second step, directionality of all other undirected edges is determined using reduction techniques (d-Separation) that can be summarized as follows. Begin by identifying indirect cause or evidential effects in the graph. That is, start by orienting any undirected edge that is part of a tuple of three vertices that have exactly one edge oriented. For example, for $v_{1} \rightarrow v_{2} - v_{3}$, orient $v_{1} \rightarrow v_{2} \rightarrow v_{3}$. Then, consider any two vertices that have an undirected edge between them. For any two such vertices, if a directed path exists through other vertices in the graph, drop the undirected edge and introduce a directed edge between the two vertices. Finally, identify any remaining common cause directionalities in the graph whereby, e.g., $v_{1} \leftarrow v_{2} \rightarrow v_{3}$. 

However, there are some problems when using PC algorithm. First, the worst case time complexity is exponential in number of vertices. This makes applications of PC Algorithm to high-dimensional datasets highly problematic. Second, the resulting DAG from PC algorithm is order-dependent. In other words, the DAG obtained from PC algorithm depends on which vertex/set of vertices is used to start the execution of algorithm. 

\subsection*{Bayesian Networks vs Regression models}

Bayesian Networks and regression models are both supervised learning techniques that can capture correlation. Regression models when they are dealing with network data are not
that flexible for the following reasons: They have properties that make it less than ideal for determining the strengths of variables in influencing outcomes and  handle worse variables that have many subtle interrelationships, not capturing them. Regression also has one broad underlying assumption that many forget: that the model is correct.

\end{document}